\documentclass[doc,floatsintext]{apa7}

\usepackage{amsmath,amssymb,amstext}
\usepackage{graphicx}
\usepackage{xcolor}
\usepackage{ulem} 
\usepackage[style=apa,sortcites=true,sorting=nyt,backend=biber, uniquename=false]{biblatex}
\title{{\bf Bayes in the Age of Intelligent Machines}}

\shorttitle{Bayes in the Age of Intelligent Machines}

\authorsnames[{1,2},1,3,1]{Thomas L. Griffiths,Jian-Qiao Zhu,\\Erin Grant,R. Thomas McCoy}

\authorsaffiliations{{Department of Computer Science,
Princeton University},{Department of Psychology, Princeton University},
{Gatsby Unit \& Sainsbury Wellcome Centre, University College London}}

\abstract{%
The success of methods based on artificial neural networks in creating intelligent machines seems like it might pose a challenge to explanations of human cognition in terms of Bayesian inference. We argue that this is not the case, and that in fact these systems offer new opportunities for Bayesian modeling. Specifically, we argue that Bayesian models of cognition and artificial neural networks lie at different levels of analysis and are complementary modeling approaches, together offering a way to understand human cognition that spans these levels. We also argue that the same perspective can be applied to intelligent machines, where a Bayesian approach may be uniquely valuable in understanding the behavior of large, opaque artificial neural networks that are trained on proprietary data.

\vspace{2mm}

}

\begin{document}

\maketitle

\begin{figure}[t!]
    \centering
    \includegraphics[width=0.8\linewidth]{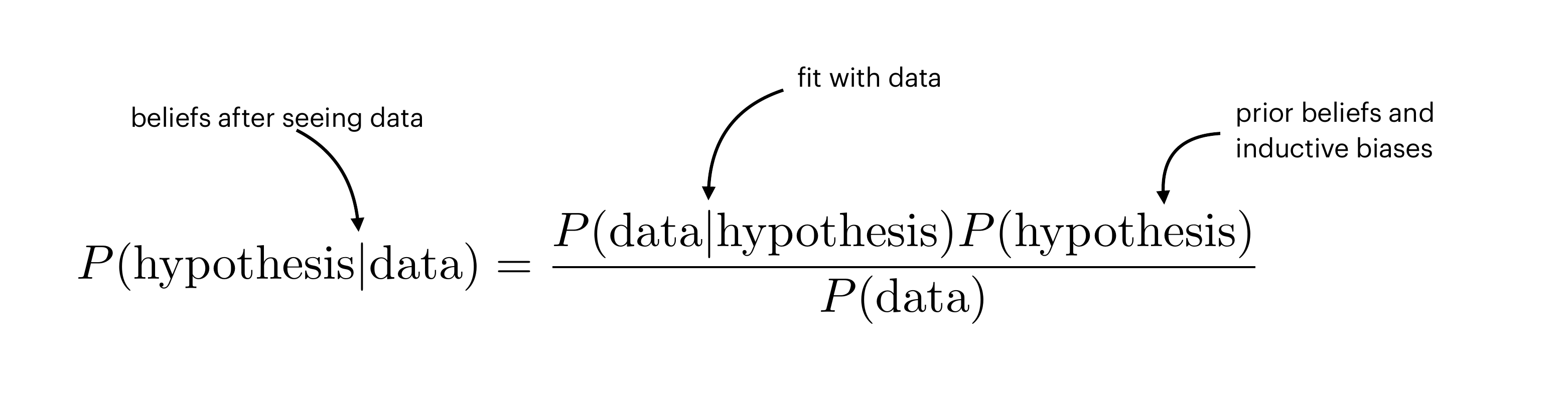}
    \caption{Bayes' rule establishes a relationship between the prior belief in a hypothesis, $P(\text{hypothesis})$, and the updated posterior belief, $P(\text{hypothesis}|\text{data})$, after incorporating evidence from data. The term $P(\text{data}|\text{hypothesis})$ represents the probability of observing the given data if the hypothesis is true, while $P(\text{data})$ acts as a normalizing constant that guarantees the resulting probabilities sum to 1.}
    \label{fig:bayes_rule}
\end{figure}

In the 18th century, the Reverend Thomas Bayes had  a radical idea: using probabilities to represent the degrees to which we believe hypotheses are true \parencite{bayes63}. He did so in the context of a gambling game: having seen some number of wins and losses, how likely are you to win? The idea of using probability theory to update our degrees of belief based on data underlies what we now call Bayes' rule (see Figure \ref{fig:bayes_rule}). Bayes  would presumably have assigned low probability to his work becoming the foundation, more than two centuries later, for Bayesian models of cognition, which explain human behavior in terms of rational belief updating (e.g., \cite{griffithsckpt10}).

Bayesian models of cognition explain inferences from limited data to an uncertain conclusion, such as inferring the meaning of a new word based on hearing that word in conversation. In Bayesian models, such inferences are framed as the result of combining data (e.g., the context in which you heard the new word) with our existing expectations about the world (e.g., expectations about what sorts of meanings a word could have). Those expectations are expressed in a ``prior distribution'' over hypotheses, with more plausible hypotheses having higher prior probability. This captures the ``inductive biases'' of a learner -- those factors other than the data that influence the hypothesis the learner selects \parencite{mitchell97}. Prior distributions can be defined over complex and expressive hypotheses, including grammars, causal structures, logical formulas, and programs
(e.g., \cite{griffithst09,goodman2011learning,rule2020child}). 

Part of the appeal of the Bayesian approach is explaining behavior via the rational solution to an abstract problem. If we accept that degrees of belief should be expressed as probabilities, then Bayes' rule solves the problem of inductive inference. This creates the opportunity to discover connections to other disciplines: statisticians and computer scientists also want to create systems that make inferences from limited data. Bayesian models of cognition benefitted from these connections, as discoveries from statistical machine learning informed accounts of human cognition (e.g., \cite{sanborngn10}). However, the last decade has seen a significant change in the landscape of machine learning. Major breakthroughs have been the result not of more sophisticated Bayesian methods, but of increasingly large artificial neural networks that are trained on increasingly large amounts of data \parencite{lecun2015deep}. Does the success of this approach in creating intelligent machines undermine the importance of Bayes' rule as a tool for understanding human cognition?\footnote{It is worth noting that there are still settings where Bayesian methods are favored in machine learning, particularly settings in which there are limited data, interpretability matters, and quantifying uncertainty is important such as science or medicine (e.g., \cite{alaa2017bayesian,padilla2021cosmological,wang2023gaussian}). Bayesian methods also underlie some of the most successful approaches to image generation (e.g., \cite{kingma2013auto,song2020score}).}

In this paper, we consider the prospects for Bayes in the age of intelligent machines. We first argue that the success of large artificial neural networks --  ``deep learning'' \parencite{lecun2015deep} -- does not pose a challenge for Bayesian models of cognition, and is actually complementary. We then argue that Bayesian models have an important new application: understanding the behavior of those intelligent machines. Deep learning has been successful in creating systems that can solve challenging problems, but the resulting systems are opaque and difficult to analyze. We suggest that the methods psychologists have developed for understanding an equally opaque and difficult-to-analyze system -- human beings -- can be adapted to make sense of large artificial neural networks, and that Bayesian models in particular have insights to offer. The key to both these arguments is the idea of levels of analysis, which we explore in more detail in the next section. 

\section{Levels of Analysis}

David Marr famously argued that information processing systems can be understood at multiple levels of analysis \parencite{Marr1982}. To borrow an analogy suggested by Marr, a biologist interested in understanding bird flight could pursue that goal in different ways. She could focus on the mechanisms underlying flight, asking {\em how} muscles, bones, and feathers translate into lift, or on the abstract principles of aerodynamics that determine {\em why} bird wings have a particular shape. When we study information processing systems -- including humans and machines -- we have the same kind of choice about the level at which we analyze those systems. 

\begin{figure}[t!]
    \centering
    \includegraphics[width=0.9\linewidth]{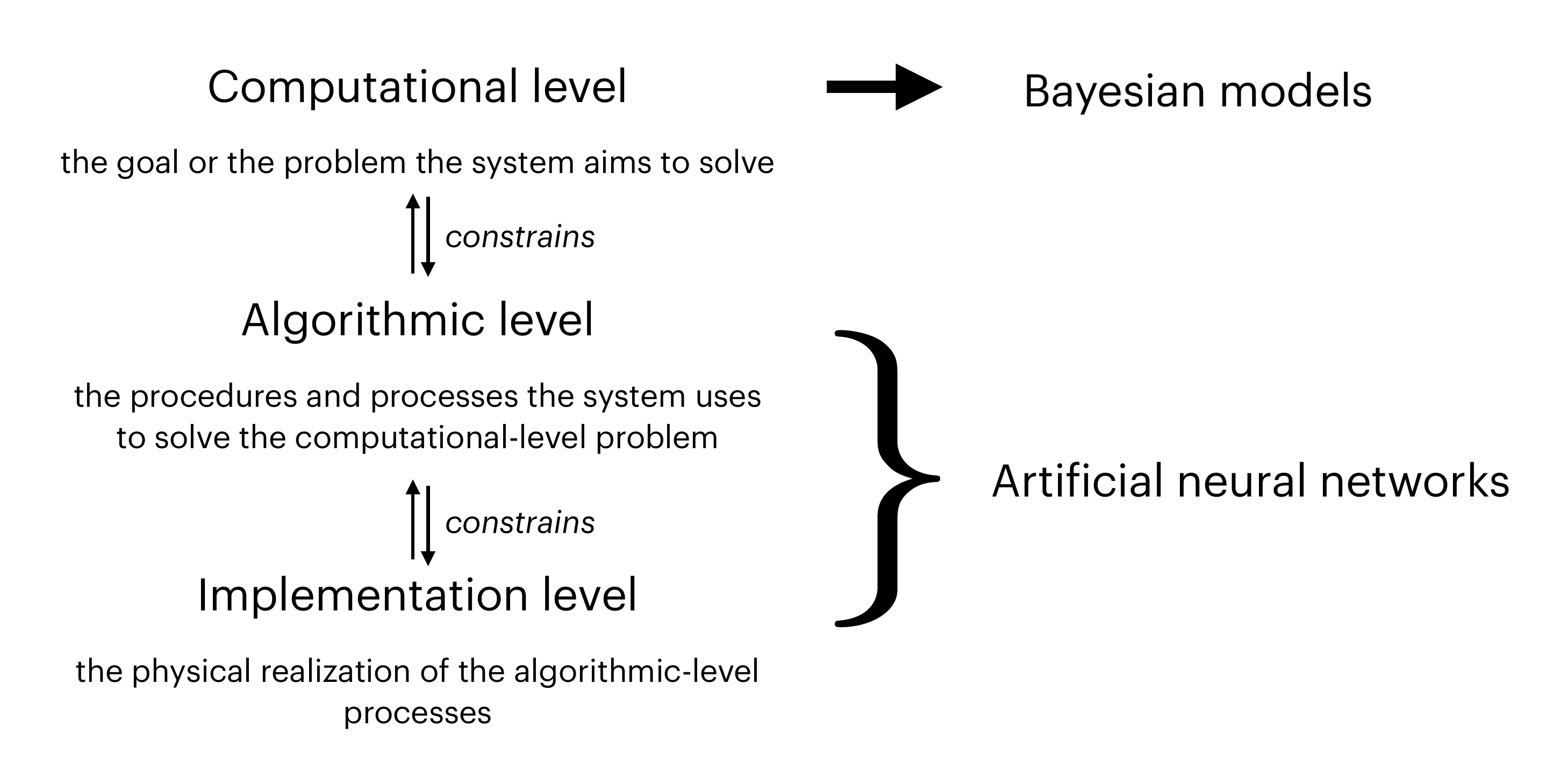}
    \caption{Marr's levels of analysis provide a framework for understanding information processing systems such as the human brain or AI systems. Different kinds of computational models engage with these different levels -- Bayesian models are typically defined at the computational level, while artificial neural networks explore hypotheses at the algorithmic and implementation levels.}
    \label{fig:marr_3_level}
\end{figure}

Marr laid out three different levels at which we can analyze an information processing system (see Figure \ref{fig:marr_3_level}). The most abstract is the {\em computational} level, where we consider the problem that the system is solving and what an ideal solution to that problem looks like. If the underlying problem involves learning or inductive inference, then an ideal solution to that problem comes from Bayes' rule -- Bayesian models of cognition are defined at this level of analysis. Next is the {\em algorithmic} level, where we consider what algorithm might (perhaps approximately) solve this problem and what representations it operates over. Finally, there is the {\em implementation} level, where we ask how that representation and algorithm can be realized physically. 

Marr also came out strongly in favor of pursuing questions at one level of analysis in particular -- the computational level. He wrote ``trying to understand perception by studying only neurons is like trying to understand bird flight by studying only feathers: It just cannot be done. In order to understand bird flight, we have to understand aerodynamics; only then do the structure of feathers and the different shapes of birds' wings make sense.'' (p.~27). Marr's argument inspired a generation of researchers pursuing computational-level analyses of cognition \parencite{shepard87,anderson90,oaksfordc94,tenenbaumg01bbs}. However, understanding human cognition is ultimately going to require answers at all three levels of analysis.

Marr's levels of analysis reveal that there are different kinds of questions that we can ask about information processing systems, each with a corresponding kind of answer, and that those answers are not necessarily in conflict with one another. For example, cognitive scientists studying human memory could offer theories at each level -- one an optimal solution to a computational problem, one a cognitive process, and one a neural circuit -- and those theories could all be correct. The key is that the theories need to be compatible: the processes at the algorithmic level must result in a reasonable approximation to the solution at the computational level, and neurons at the implementation level must, in turn, execute something like that algorithm.

\section{Bayes and Deep Learning are Complementary Approaches}

Having introduced the idea of levels of analysis, we can now make our first argument: the success of deep learning is not a challenge to Bayesian models of cognition because these two approaches address different levels of analysis and are compatible. The first of these claims is relatively straightforward: As noted above, Bayesian models of cognition are explicitly defined at the computational level. By contrast, accounts of human cognition based on artificial neural networks typically situate themselves at the algorithm or implementation levels, focusing on cognitive or neural processes rather than abstract problems and their ideal solution (e.g., \cite{mcclellandbnprss10}). The key issue is thus whether these approaches are compatible. 

There are both theoretical arguments and empirical results supporting the view that they are compatible. As noted above, Bayes' rule is an optimal solution to problems of inductive inference assuming that the world is well-described by a particular prior distribution over hypotheses. Artificial neural networks are trained to minimize a loss function (also known as an error function or an objective function), and those functions have natural probabilistic interpretations. For example, minimizing the cross-entropy loss corresponds to maximizing the probability of discrete data under a Bernoulli or multinomial distribution, and minimizing the squared-error loss corresponds to maximizing the probability of continuous data under a Gaussian distribution. Artificial neural networks thus have a direct interpretation as a kind of probabilistic model, and should be seeking a hypothesis (i.e., a set of connection weights) that assigns high probability to the observed data. We thus need to show that these models capture the impact of the prior distribution on selecting that hypothesis in a way that is consistent with Bayesian inference. 

Classic theoretical results show that the algorithms used to train artificial neural networks -- gradient descent with weight decay \parencite{mackay95} or early stopping \parencite{bishop1995regularization} -- are consistent with imposing a Gaussian prior on the weights of the network. Other analyses show how specific neural network architectures can be used to implement Bayesian inference for arbitrary prior distributions \parencite{shig09}. More recently, researchers have developed methods for ``amortizing'' Bayesian inference using neural networks, by explicitly training networks to approximate the output of Bayes' rule for a specific prior distribution  \parencite{dasgupta2021memory}. Finally, empirical analyses of deep learning models show that they seem to internalize information that can be used to perfectly reconstruct relevant Bayesian posterior distributions \parencite{mikulik2020meta}.
Even if the strategies that deep learning systems are employing are not consistent with Bayes with any particular prior, a Bayesian ideal is at least a starting point in making sense of these systems (e.g., \cite{raventos2023pretraining}). Systematic deviations from that Bayesian ideal would suggest the behavior could be explained as approximate Bayesian inference under resource constraints, an idea that has also been used in understanding human cognition \parencite{griffiths2015rational}.

But what about the grammars, causal structures, logical formulas, and programs that play a prominent role in Bayesian models of cognition? There is nothing like those built into artificial neural networks, which represent the world with continuous weights and activations. This is problematic if the structured representations used in defining Bayesian models are interpreted as a claim about the algorithmic and implementation levels, which are the levels that artificial neural networks are designed to model.

However, it is not problematic if we view those representations as being primarily useful in providing a way to specify human-like inductive biases. In this case, the question becomes whether artificial neural networks can manifest inductive biases consistent with those specified using such structured representations.

\begin{figure}[t]
    \centering
    \includegraphics[width=\textwidth]{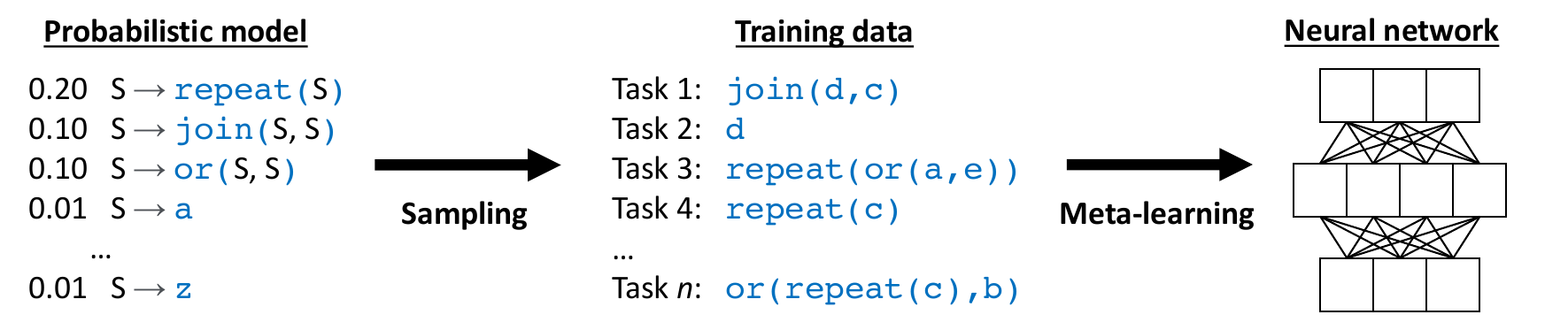}
    \caption{Distilling a Bayesian model's prior into a neural network. We first define a prior using a probabilistic model (left). Then, we sample many tasks from that prior (middle). Here, each ``task'' is a formal rule defining a set of strings; e.g., \textit{repeat(c)} defines the set containing \textit{c}, \textit{cc}, \textit{ccc}, etc. Finally, meta-learning is used to create a neural network that is trained to perform those tasks (right), giving it a prior that approximates the one we started with. Meta-learning is a process in which a learner encounters many different tasks and leverages the commonalities across those tasks to gain inductive biases that enable it to learn new tasks more easily. 
    }
    \label{fig:distillation}
\end{figure}

Our recent work suggests that this is the case. In previous work, we had shown that meta-learning -- a procedure in which the initial weights of a neural network are adapted to make it easier for that network to perform a variety of tasks -- can be interpreted as learning an appropriate Bayesian prior distribution for those tasks \parencite{grant2018recasting}. We have built upon that to develop a method for ``distilling'' an explicit prior distribution from a Bayesian model into a neural network (\cite{mccoy2023modeling}; see Figure \ref{fig:distillation}). This method generates a set of tasks by sampling from that prior distribution, then uses meta-learning to create a neural network that is easily adapted to perform those tasks. We have shown that this approach can distill an abstract prior distribution over formal languages -- itself defined by a grammar -- into a set of weights for a recurrent neural network. We anticipate that a similar approach will be effective for other structured prior distributions, providing the last piece of evidence that deep learning and Bayesian models are complementary approaches to understanding the mind. 

\section{From Modeling People to Modeling Machines}

The claim that Bayesian models of cognition and deep learning are complementary has a broader implication: that we can expand the scope of Bayesian modeling from humans to machines. Deep learning has created systems that can solve challenging problems, but it has a number of limitations. Deep neural networks are opaque and hard to interpret, particularly when their internal workings are withheld by the companies that create them. This leaves computer scientists in the unfamiliar territory of trying to make sense of complex information processing systems via their behavior. This, of course, is a problem that psychologists are intimately familiar with, and a place where Bayesian models of cognition might be uniquely helpful.

\begin{figure}[t!]
    \centering
    \includegraphics[width=0.95\textwidth]{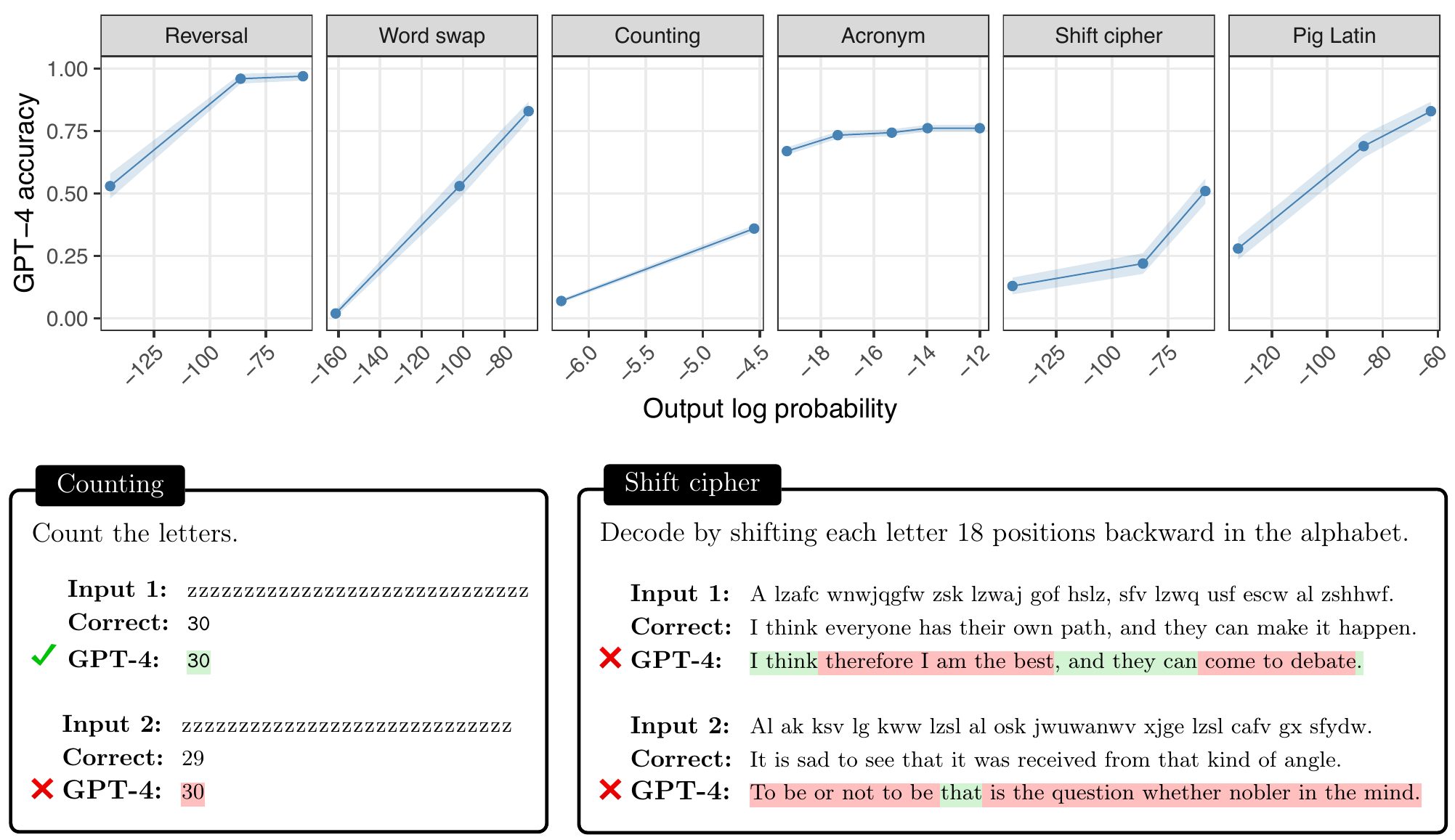}
    \caption{As predicted by a Bayesian analysis, GPT-4 performs better when the correct output is high-probability than when it is low-probability, even when it is applied to deterministic tasks (reversing a list, swapping certain words in a sequence, counting letters, forming acronyms, decoding a simple cipher, and converting English into Pig Latin). For instance, when counting letters (bottom left), GPT-4 is more likely to answer correctly when the answer is 30 than when it is 29; in natural text, the number 30 occurs over four times as often as 29. Another sign of GPT-4's probability sensitivity is a tendency to erroneously produce high-probability sequences such as well-known quotations (``I think therefore I am'' and ``To be or not to be'') (bottom right).}
    \label{fig:embers}
\end{figure}

We can apply Marr's levels of analysis to AI systems. This idea is novel in the context of machine learning, where a practitioner might just think about  choosing a method -- Bayes or deep learning -- to use to solve an engineering problem. But the compatibility of these approaches means that even if the practitioner chooses to build a deep learning system, we can still ask what an ideal solution to that problem looks like and use that solution to understand the system they have built. Bayesian models are particularly valuable in this setting because they specify such an ideal solution, which we can use for understanding the behavior of information processing systems even if the underlying representations and algorithms differ from Bayesian inference.

As a simple illustration, \textcite{li2023gaussian} showed it is possible to model simple artificial neural networks using Gaussian processes, a Bayesian formalism that has also been used to capture aspects of human function learning \parencite{lucas2015rational}. The inductive biases of these networks are thus described by specific prior distributions over functions, clarifying their underlying assumptions and allowing easy comparison to the inductive biases of human learners. 

The Bayesian approach can also provide insight into the behavior of large, complex artificial neural networks. McCoy et al.~(2023) used this perspective to analyze the performance of GPT-4, a popular large language model. A Bayesian analysis suggests that how well this model performs will be influenced by the probability with which the required answer appears in the text used to train the model, and this prediction is borne out in some surprising ways that the model behaves even when solving deterministic problems (see Figure~\ref{fig:embers}).

\nocite{mccoy2023embers}

\begin{figure}[t!]
    \centering
    \includegraphics[width=0.95\linewidth]{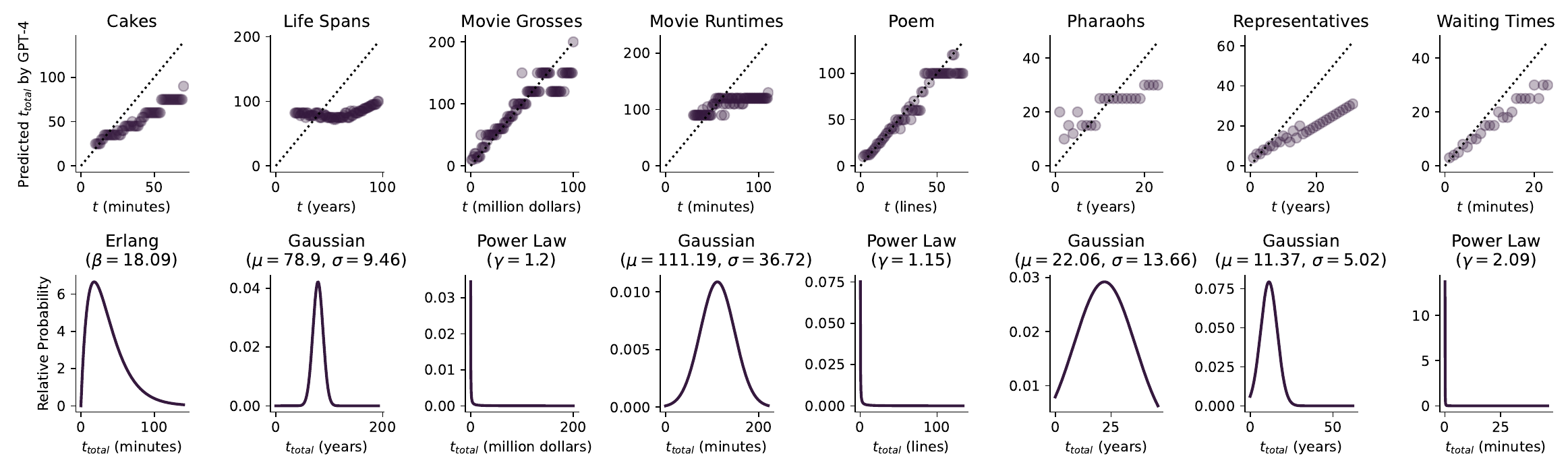}
    \caption{GPT-4's forecasts for different everyday events (left to right: baking times for cakes, life spans, movie grosses, movie run times, poem lengths, reigns of Egyptian Pharaohs, terms of members of the U.S. House of Representatives, waiting times for calling a box office). The top row of plots illustrates GPT-4's predictions of total duration or extent, $t_{total}$, for each sample event duration $t$ (dots). Dashed lines represent predictions using a fixed non-informative prior, $P(t_{total}) \propto 1/t_{total}$, which predicts $t_{total} = 2t$. The bottom row presents the recovered prior distributions of $t_{total}$ for each event based on GPT-4's predictions, using  a Bayesian model presented in \textcite{griffithst06future}. $\beta$ represents the scale of an Erlang distribution.  $\mu$ is the mean and $\sigma$ is the standard deviation in a Gaussian distribution. $\gamma$ indicates power in a power-law distribution.}
    \label{fig:gpt_everyday_cognition}
\end{figure}

Finally, preliminary results suggest that a more explicit version of Bayesian models of cognition can be applied to AI systems. For example, \textcite{griffithst06future} used a ``predicting the future'' task to explore human prior distributions for various everyday events. The same task can be given to a large language model, and a Bayesian model can be used to infer the prior distribution that implicitly informs the responses. Applying this approach shows that the model has accurately internalized a variety of everyday distributions in a way that is similar to humans (see Figure \ref{fig:gpt_everyday_cognition} and Appendix). 

\section{Conclusion}

The success of artificial neural networks in machine learning would seem to pose a challenge to Bayesian models of cognition. Instead, we argue, it presents an opportunity. First, it provides a way to begin to explain human behavior at multiple levels of analysis, with Bayesian models at the computational level and neural networks at the algorithmic and implementation levels. Second, the artificial neural networks used in machine learning are opaque, complex, and difficult to interpret -- just like humans. Bayesian models thus provide a new tool for exploring the inductive biases of machines. To echo Marr, it seems unlikely that we will understand cognition by studying only artificial neurons.

\newpage

\section{Acknowledgments}

This research project and related results were made possible with the support of the NOMIS Foundation and the NSF SPRF program (grant no. 2204152).

\printbibliography

\newpage
\section{Appendix}
\subsection{Methods}
To demonstrate how Bayesian models can be used to reveal the prior distributions implicitly assumed by AI systems based on deep learning, we had GPT-4 undertake the “predicting the future” task from \textcite{griffithst06future}. The goal of the task was to forecast a total duration or quantity $t_{\mathrm{total}}$ based on an observed partial duration or quantity $t$. As in the original study, we systematically varied the value of $t$ (details in the Prompts section). However, since we used GPT-4, we could more densely sample $t$ values. 

\subsection{Prompts}
\subsubsection{Cakes}
\texttt{"Each of the questions below asks you to predict something, either a duration or a quantity, based on a single piece of information. Please read each question and respond only with your prediction on the line below it. We're interested in your intuitions, so please don't make complicated calculations; just tell us what you think! Imagine you are in somebody's kitchen and notice that a cake is in the oven. The timer shows that it has been baking for {t} minutes. What would you predict for the total amount of times the cake needs to bake? Predicted\_number\_of\_minutes=”}
where $t$ represents integers ranging from 10 to 70.

\subsubsection{Life Spans}
\texttt{“Each of the questions below asks you to predict something, either a duration or a quantity, based on a single piece of information. Please read each question and respond only with your prediction on the line below it. We're interested in your intuitions, so please don't make complicated calculations; just tell us what you think! Imagine you hear about a movie that has taken in {t} million dollars at the box office, but don't know how long it has been running. What would you predict for the total amount of box office intake for that movie? Predicted\_number\_of\_million\_dollars=”}
where $t$ represents integers ranging from 1 to 100.

\subsubsection{Movie Grosses}
\texttt{“Each of the questions below asks you to predict something, either a duration or a quantity, based on a single piece of information. Please read each question and respond only with your prediction on the line below it. We're interested in your intuitions, so please don't make complicated calculations; just tell us what you think! Imagine you hear about a movie that has taken in {t} million dollars at the box office, but don't know how long it has been running. What would you predict for the total amount of box office intake for that movie? Predicted\_number\_of\_million\_dollars=”}
where $t$ represents integers ranging from 1 to 100.

\subsubsection{Movie Runtimes}
\texttt{“Each of the questions below asks you to predict something, either a duration or a quantity, based on a single piece of information. Please read each question and respond only with your prediction on the line below it. We're interested in your intuitions, so please don't make complicated calculations; just tell us what you think! If you made a surprise visit to a friend, and found that they had been watching a movie for {t} minutes, what would you predict for the length of the movie? Predicted\_number\_of\_minutes=”
}
where $t$ represents integers ranging from 30 to 110.

\subsubsection{Poem}
\texttt{“Each of the questions below asks you to predict something, either a duration or a quantity, based on a single piece of information. Please read each question and respond only with your prediction on the line below it. We're interested in your intuitions, so please don't make complicated calculations; just tell us what you think! If your friend read you her favorite line of poetry, and told you it was line {t} of a poem, what would you predict for the total length of the poem? Predicted\_number\_of\_lines=”}
where $t$ represents integers ranging from 2 to 67.

\subsubsection{Paraohs}
\texttt{“Each of the questions below asks you to predict something, either a duration or a quantity, based on a single piece of information. Please read each question and respond only with your prediction on the line below it. We're interested in your intuitions, so please don't make complicated calculations; just tell us what you think! If you opened a book about the history of ancient Egypt to a page listing the reigns of the pharaohs, and noticed that at 4000 BC a particular pharaoh had been ruling for {t} years, what would you predict for the total duration of his reign? Predicted\_number\_of\_years=”}
where $t$ represents integers ranging from 1 to 23.

\subsubsection{Representatives}
\texttt{“Each of the questions below asks you to predict something, either a duration or a quantity, based on a single piece of information. Please read each question and respond only with your prediction on the line below it. We're interested in your intuitions, so please don't make complicated calculations; just tell us what you think! If you heard a member of the U.S. House of Representative had served for {t} years, what would you predict his total in the House would be? Predicted\_number\_of\_years=”}
where $t$ represents integers ranging from 1 to 31.

\subsubsection{Waiting Times}
\texttt{“Each of the questions below asks you to predict something, either a duration or a quantity, based on a single piece of information. Please read each question and respond only with your prediction on the line below it. We're interested in your intuitions, so please don't make complicated calculations; just tell us what you think! If you were calling a telephone box office to book tickets and had been on hold for {t} minutes, what would you predict for the total time you would be on hold? Predicted\_number\_of\_minutes=”}
where $t$ represents integers ranging from 1 to 23.

\subsection{Analysis}
Our analysis of GPT-4’s responses paralleled the approach used for human data (detailed in the Appendix of \cite{griffithst06future}). We adopted two key assumptions to recover prior distributions from GPT-4’s responses. First, we considered GPT-4’s estimate of $t_{\mathrm{total}}$, labeled $t^\ast$, as the posterior median, meaning that $P(t_{\mathrm{total}}>t^\ast|t)=0.5$. Second, we assumed that potential prior distributions include power-law, Erlang, and Gaussian forms.

For the power-law and Erlang distributions, analytically solvable prediction functions can be obtained for $P(t_{\mathrm{total}}>t^\ast|t)=0.5$. Specifically, for a power-law prior with parameter $\gamma$, $P(t_{\mathrm{total}})\propto{t_{\mathrm{total}}}^{-\gamma}$, the prediction function has the following form $t^\ast=2^{1/\gamma}t$; for an Erlang prior with scale parameter $\beta$, $P(t_{\mathrm{total}})\propto t_{\mathrm{total}} \text{exp}(-t_{\mathrm{total}}/\beta)$, the prediction function has the following form $t^\ast=t+\beta \log 2$. For the Gaussian prior, characterized by mean $\mu$ and standard deviation $\sigma$, we used numerical methods to approximate $t^\ast$ values.

We fitted prediction functions for power-law, Erlang, and Gaussian distributions to align with GPT-4’s responses, aiming to minimize the mean squared differences between the GPT-4’s responses and the predicted values of $t_{\mathrm{total}}$. As shown in Figure 5, the power-law priors with $\gamma=1.20\ \text{and}\ 1.15$ best matched GPT-4’s responses for movie grosses and poem lengths, respectively. The Erlang priors with $\beta=18.09$ provided the best fit for cake baking times. The Gaussian priors with $\mu=78.90,\ 111.19,\ \ 22.06,\ 11.37$ and $\sigma=9.46,\ 36.72,\ 13.66,\ 5.02$ was most effective for human life spans, movie runtimes, reigns of pharaohs, and terms of U.S. representatives, respectively.

\end{document}